\documentclass{book}    
\usepackage{piers}  
\usepackage{cite}
\begin{document}
	
	\title{Categorical Difference and Related Brain Regions of the Attentional Blink Effect}
	\maketitle
	
	\author      {Xiaohong Ji}
	\affiliation {Tongji University}
	\address     {}
	\city        {Shanghai}
	\postalcode  {}
	\country     {China}
	\phone       {18967640777}    
	\fax         {}    
	\email       {2030695@tongji.edu.cn}  
	\misc        { }  
	\nomakeauthor
	
	\author      {Renzhou Gui}
	\affiliation {Tongji University}
	\address     {}
	\city        {Shanghai}
	\postalcode  {}
	\country     {China}
	\phone       {13472419478}    
	\fax         {}    
	\email       {rzgui@tongji.edu.cn}  
	\misc        { }  
	\nomakeauthor
	
	\begin{authors}
		
		{\bf Renzhou Gui and Xiaohong Ji}\\
		\medskip
        Tongji University, China
		
	\end{authors}
	
	\begin{paper}
		
		\begin{piersabstract}
Attentional blink (AB) is a biological effect, showing that for 200 to 500ms after paying attention to one visual target, it is difficult to notice another target that appears next, and attentional blink magnitude (ABM) is a indicating parameter to measure the degree of this effect. Researchers have shown that different categories of images can access the consciousness of human mind differently, and produce different ranges of ABM values. So in this paper, we compare two different types of images, categorized as animal and object, by predicting ABM values directly from image features extracted from convolutional neural network (CNN), and indirectly from functional magnetic resonance imaging (fMRI) data. First, for two sets of images, we separately extract their average features from layers of Alexnet, a classic model of CNN, then input the features into a trained linear regression model to predict ABM values, and we find higher-level instead of lower-level image features determine the categorical difference in AB effect, and mid-level image features predict ABM values more correctly than low-level and high-level image features. Then we employ fMRI data from different brain regions collected when the subjects viewed 50 test images to predict ABM values, and conclude that brain regions covering relatively broader areas, like LVC, HVC and VC, perform better than other smaller brain regions,  which means AB effect is more related to synthetic impact of several visual brain regions than only one particular visual regions.
		\end{piersabstract}
		

\psection{Introduction}

Human brain is bombarded with an ocean of sensory information from the outside world at every moment, however, not all of the information it receives can be fully analyzed. Due to the insufficient processing power to consciously perceive the incoming sensory infomation, the brain develops the attentional mechanisms to select relatively more important aspects of the sensory world, thus is characterized by the bottlenecks that limit what we can perceive and act in multitask settings\cite{1}. \par

Particularly, for simultaneous visual events, they often have to compete to be processed by the brain. The attentional blink (AB) paradigm is one strong case of visual attentional competition\cite{2}. In this paradigm, two visual objects, T1 and T2, respectively the first and second targets, appear successively in a rapid serial visual presentation (RSVP) with other distractive items. Experiments show that when T2 appears within 200 to 500ms after T1, the observer often fails to detect T2 correctly\cite{3}. However, when the observer ignores T1, this phenomenon will disappear. The explanation of attentional blink effect is usually attentional rather than perceptual, so attentional blink is also considered as a defect, reflecting the limited capacity of information that the human brain can process and the mechanism of information selection. \par

As Figure \ref{fig4} shows, in a typical AB paradigm, subjects are presented with a RSVP consisting of distractors and two embedded targets, T1 and T2. When the time interval bewteen T1 and T2 is 200ms and 800ms, it is called lag2 and lag8 situations respectively. 
In this work, AB magnitudes (ABM) are the  difference in the probability of correctly identifying T2 between lag 8 and lag 2, which can act as an indicator for the level of AB effect.\par

\begin{figure}[htbp]
	\vspace{-5 pt}
	\centerline{\includegraphics[width=8cm]{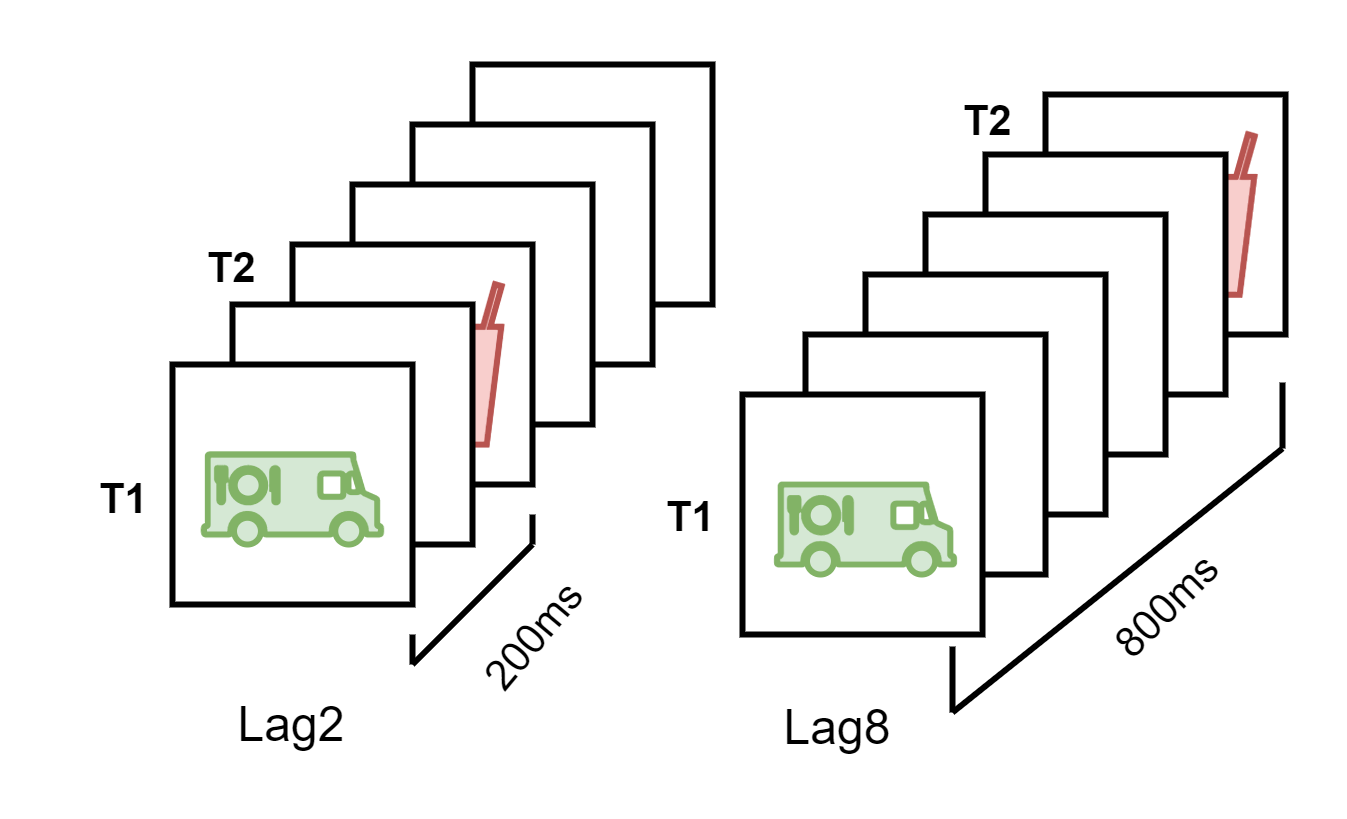}}
	\caption{Attentional blink paradigm with lag2 and lag8 situations}
	\label{fig4}
	\vspace{-5 pt}
\end{figure}


Many studies have found specific brain regions that are related to AB effect using event-related functional magnetic resonance imaging (fMRI), in which subjects are required to view a rapid stream of visual pictures and T2 is either presented within or outside the AB  period, or not at all. Right intraparietal, posterior parietal, medial prefrontal, frontal and occipitotemporal cortex activations are found predominantly associated with the AB. On the other hand, activation of the anterior cingulate, inferotemporal and cerebellumwere are also involved in the conscious perception of T2 in an AB paradigm\cite{4,5,6}. Among these researches, it is worth noting that primary visual cortex (V1) is also found linked to AB effect and conscious vision, for the retinotopic organization of V1 will result in different encoding for T1 and T2, thus affect how they enter the consciousness\cite{7}. However, other visual cortices, like V2, V3 and V4 have rarely been explored in AB paradigm, so in this work we experiment with these brain regions related to visions about their linkage with AB effect.

On the other hand, different objects may be processed in the brain throughout the visual stream in different ways. And there are studies showing that there are indeed categorical difference in AB effect, that is, for images from different classes, for example, animate and inanimate images, the level of AB effect can also be distinguishingly different. Besides, animate objects are more often consciously perceived than inanimate objects, thus show lower level of AB effect in previous studies\cite{8,9}. For narrower classification, more specifically, difference of some sub-classes like fruits and vegetables, processed foods, objects, scenes, animal bodies, animal faces, human bodies and human faces have been explored in \cite{10}.

In this paper, we focus on the categorical difference and related brain regions of AB effect. We first propose a two-stage model for predicting ABM values, with fMRI data. Then for categorical difference, we compare two sets of images from two sub-categories, namely animal and object classes with regard to their ABM values. For related brain regions, we compare 10 brain regions related to visual systems, namely V1–V4, the lateral occipital complex (LOC), fusiform face area (FFA), parahippocampal place area (PPA), lower visual cortex (LVC, covering V1–V3), higher visual cortex (HVC, covering regions around LOC, FFA and PPA) and the entire visual cortex (VC,  covering all of the visual subareas listed above), with regard to their correlation with AB effect. It is assumed that (1) for target images of the same category, image features predicted by fMRI are similar. (2) for the same category of target images, the ABM values predicted by image features are also similar. (3) the individual difference of subjects have little influence in this study.

\psection{Model Structure and Training}

First of all, inspired by \cite{10} and \cite{11}, we propose a two-stage model for predicting ABM values with fMRI data for every particular image. In the first stage, fMRI data collected while subjects were viewing categorical images are correlated with visual image features, which are extracted by a typical convolutional neural network, Alexnet. We use linear regression model to establish the relationship between fMRI data and image features. Then in the second stage, we use another linear regression model to establish the relationship between image features and ABM values. So as a whole this model can predict ABM values from fMRI data. The model framework is shown in Figure \ref{fig1}.\par

It can be seen that image features serve as a link between fMRI data and ABM values, so how to extract image features and which layer of image features to extract are crucial to the prediction of ABM results from fMRI data. We choose a classic convolutional neural network (CNN), Alexnet, proposed by Alex Krizhevsky et.al in 2012\cite{12}, to extract image features. Alexnet model is realized by the MindSpore Lite tool\cite{13}, and pretrained on CIFAR-10 dataset. Alexnet consists of 8 layers, the first five layers are the convolution layers, and the last three layers are the full connection layers. The dimensions of the convolution layers are: the first layer (conv1) is 96$\times$55$\times$55, the second layer (conv2) is 256$\times$27$\times$27, the third layer (conv3) is 384$\times$13$\times$13, and the fourth layer (conv4) is 384$\times$13$\times$13, the fifth layer (conv5) is 256$\times$13 $\times$13. Fully connected layers fc6, fc7 and fc8 are composed of one-dimensional arrays with 4096, 4096 and 10 units respectively. Since fc8 is the output layer, we ignore it in image feature extraction process.
 
For the first stage model training, there are 1200 training images, the corresponding fMRI data from 10 brain regions are taken as the input of the linear regression model, and features from 7 layers of Alexnet (except the last output layer) are taken as the labels, thus enables the training of the linear regression model. For the second stage model training, there are 41 training images, the corresponding image features are the input to the linear regression model, and the ABM values are the labels of the linear regression model. Cross validation is used to train the linear regression model. Thus, the parameters of two linear regression models in Figure  \ref{fig1} are determined.\par

\begin{figure}[htbp]
	\vspace{-5 pt}
	\centerline{\includegraphics[width=14.5cm]{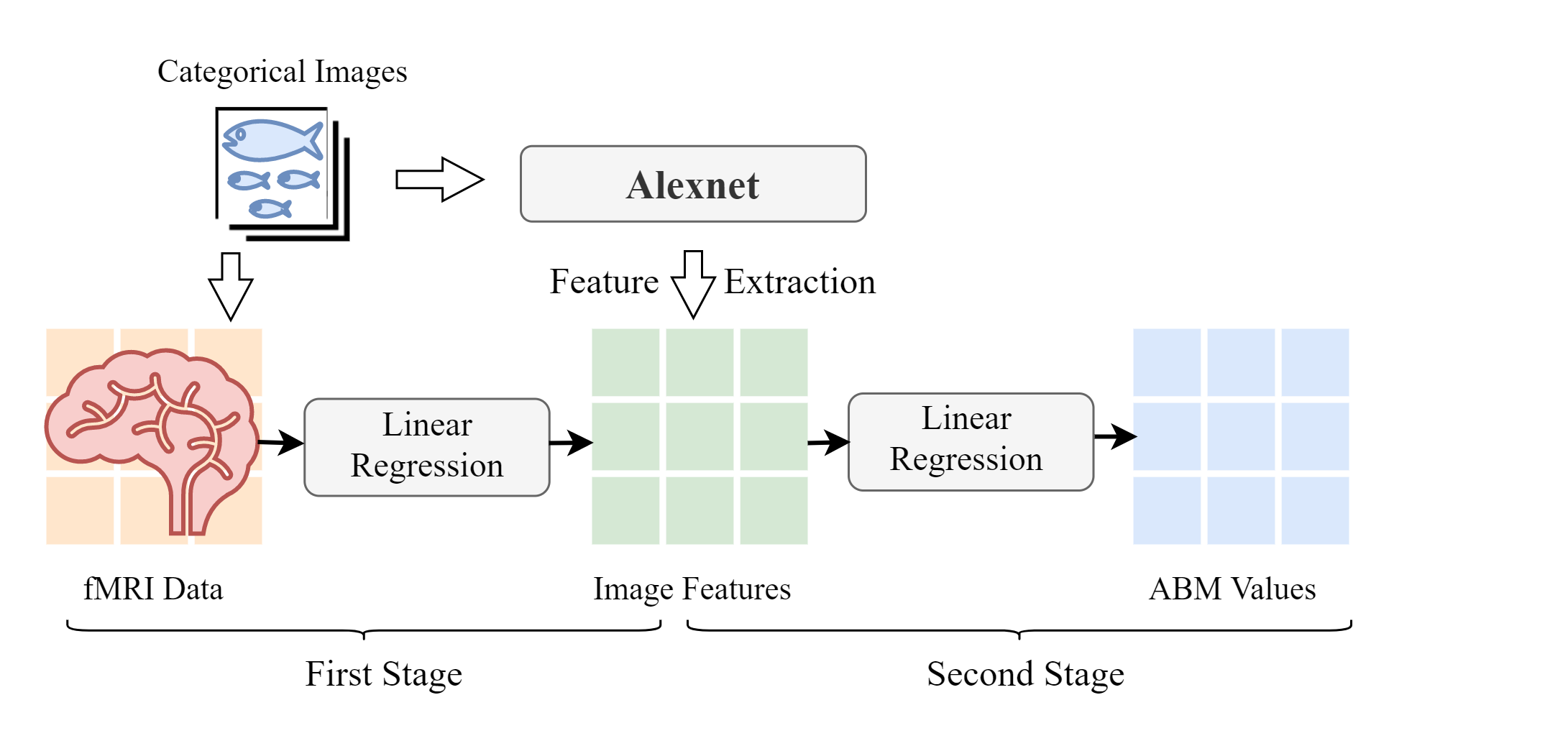}}
	\caption{Model framework}
	\label{fig1}
	\vspace{-5 pt}
\end{figure}

\psection{Categorical Difference in the Attentional Blink Effect}

Among the 50 test images, we manually choose 12 images as the animal category and 12 images as the object category, and extract the average image features of these two categories from seven layers of Alexnet respectively, then input them separately to the second trained linear regression model, thus get predicted ABM values for animal and object categories for seven CNN layers, as shown in Figure \ref{fig5}.\par

It shows that for image features extracted from conv1 and conv2, the predicted ABM values for animal category are close to object category, while for high layers of CNN layers the difference increases between the two categories, indicating that low-level image features may not be a distinction between the two categories, and the two categories may share low-level scene statistics. Thus higher-level features are the reason for category difference that distinguish two categories. On the other hand, for conv3 and conv5, predicted ABM values for animal category are lower that that of object category, which is consistent with previous research with conclusions that animate targets are more often consciously perceived in RSVP \cite{8}. However, for conv4, fc6 and fc7, the situation is the opposite. This suggests that mid-level image features may predict ABM values more correctly and more consistent with the real world empirical facts.\par

As a result, in this part, we conclude that higher-level instead of lower-level image features determine the categorical difference in AB effect, and mid-level image features predict ABM values more correctly than low-level and high-level image features, thus the magnitude of AB effect is most closely related to mid-level T2 image features.

\begin{figure}[htbp]
	\vspace{-5 pt}
	\centerline{\includegraphics[width=9cm]{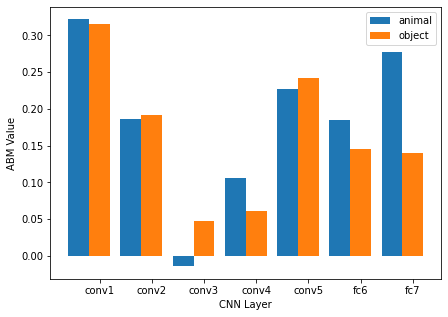}}
	\caption{Comparison of ABM values between animal and object categories for each CNN layer}
	\label{fig5}
	\vspace{-5 pt}
\end{figure}

\psection{Related Brain Regions of the Attentional Blink Effect}

In this part, we try to find out the related brain regions in the following way: first of all, for the 50 test images, CNN features of each single image are extracted respectively, then input to the second trained linear regression model, so predicted ABM values for each image and each CNN layer are obtained, as shown in Figure \ref{fig2}. As a result, these 350 ABM values are predicted directly from 50 test image features. The x axis represents the seven layers of CNN, with "1" refers to conv1, "6" refers to fc6, and so on. And the y axis represents the number of 50 test images, with "1" refers to the image numbered as NO.1, and so on. Lastly, the z axis represents ABM values. Next, we will take these 350 predicted ABM values as the "reference" values, which means they are predicted directly from image features. We can see different test images have very different predicted ABM values (look along the y axis), and for a particular image, ABM values can also vary with CNN layers (look along the x axis). We average the 50 predicted ABM values for each CNN layer, as shown in Figure \ref{fig6}. Notably, ABM values fo conv3 and conv5 are slightly higher than other layers, which are consistent with conclusions drawn in the last part of this work.\par
		
Then, for the 50 test images, the corresponding 50 sets of fMRI data are input to the fisrt trained linear regression model, getting the predicted image features from fMRI data, which are then input to the second trained linear regression model, and finally ABM values are predicted, as shown in Figure \ref{fig7}. So compared with the "reference" ABM values mentioned in the last paragraph, these 350 ABM values are predicted indirectly from fMRI data, with image features as internal transitions. With these two sets of ABM values predicted directly from image features and indirectly from fMRI data, we calculate the mean square error (MSE) between them, as Figure \ref{fig3} shows. We can see that MSE for LVC, HVC and VC regions are relatively lower than other brain regions, indicating that brain regions covering relatively broader areas perform better than those smaller brain regions tested in this experiment, which means AB effect is more related to synthetic impact of several visual brain regions than one particular visual regions.\par	
		
\begin{figure}[htbp]
	\vspace{-5 pt}
	\centerline{\includegraphics[width=8.5cm]{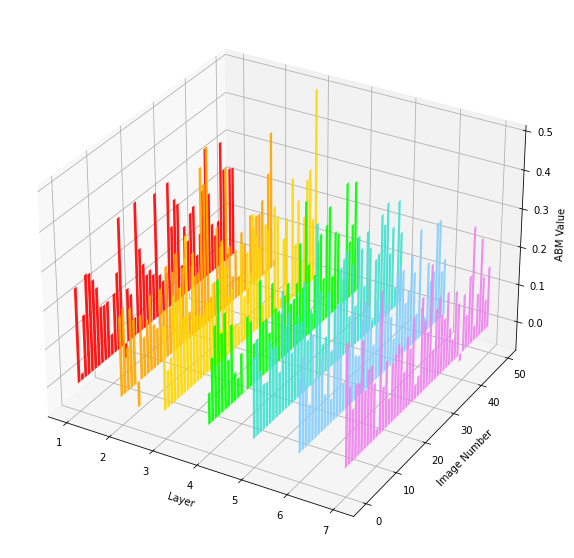}}
	\caption{ABM values predicted directly by 50 test image features for 7 CNN layers}
	\label{fig2}
	\vspace{-5 pt}
\end{figure}

\begin{figure}[htbp]
	\vspace{-5 pt}
	\centerline{\includegraphics[width=8cm]{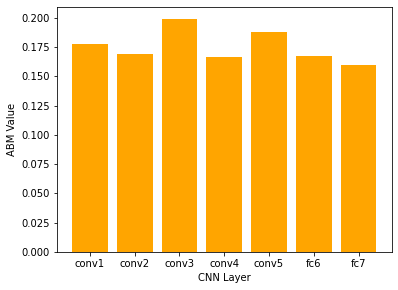}}
	\caption{Average ABM values for each CNN layer}
	\label{fig6}
	\vspace{-5 pt}
\end{figure}

\begin{figure}[htbp]
	\vspace{-5 pt}
	\centerline{\includegraphics[width=8.5cm]{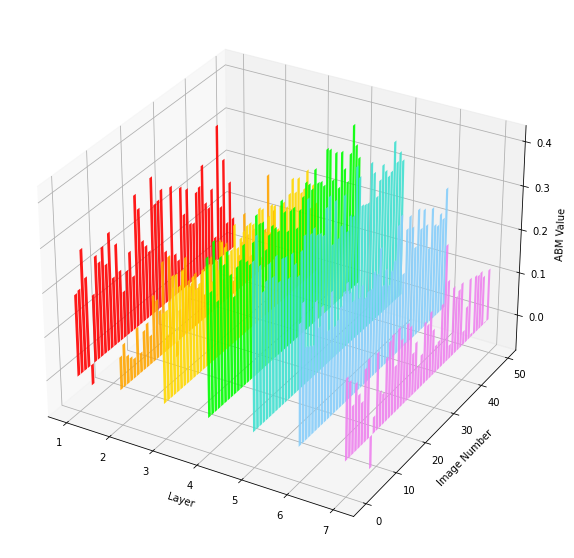}}
	\caption{ABM values predicted indirectly by fMRI data corresponding to 50 test 
		images for 7 CNN layers}
	\label{fig7}
	\vspace{-5 pt}
\end{figure}

\begin{figure}[htbp]
	\vspace{-5 pt}
	\centerline{\includegraphics[width=8cm]{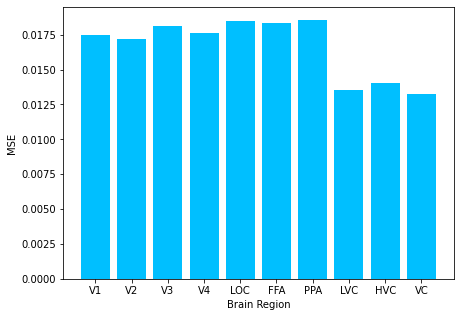}}
	\caption{Comparison of MSE for 10 visual brain regions}
	\label{fig3}
	\vspace{-5 pt}
\end{figure}

\psection{Conclusion}

In this paper, we focus on two aspects of AB effect, categorical difference and related brain regions. First we establish a two-stage model with two linear regression models, and train them separately with existing datasets. Then, for categorical difference, we compare two sets of images categorized as animal and object, by predicting ABM values from image features extracted from each CNN layer. We find that higher-level instead of lower-level image features determine the categorical difference in AB effect, and mid-level image features predict ABM values more correctly that low-level and high-level image features, thus the magnitude of AB effect is most closely related to the mid-level T2 image features. On the other hand, for related brain regions, we compare two sets of ABM values, respectively predicted directly from extracted image features and indirectly from fMRI data, and compute the MSE of these two sets of ABM values. We conclude that brain regions covering relatively broader areas perform better than those smaller brain regions tested in this experiment, which means AB effect is more related to synthetic impact of several visual brain regions than one particular visual regions. In future reasearch, we hope we can explore difference in AB effect for more image subcategories, like scene, human faces and so on. Also, other brain regions related to AB effect besides these visual regions mentioned in this work can also be experimented.

\ack

We thank MindSpore for the partial support of this work, which is a new deep learning 
computing framework\footnote{https://www.mindspore.cn/}. This work is also supported by Special Fund for Basic Research on Scientific Instruments of the National Natural Science Foundation of China via the Grant No.4182780021, Taihang mountain highway Program and PowChina Hebei Transportation Highway Investment Development Co., Ltd via the Grant No.TH-201908, emeishan-hanyuan highway Program and Sichuan Railway Investment Group Co., Ltd via the Grant No.SRIG2019GG0004, and Department of Science and Technology of Guizhou Province via the Grant No.[2018]3011.

\end{paper}
	
\end{document}